# MACHINE LEARNING OF PHONOLOGICALLY CONDITIONED NOUN DECLENSIONS FOR TAMIL MORPHOLOGICAL GENERATORS

K.Rajan [1], Dr.V.Ramalingam [2], Dr.M.Ganesan [3]

[1] Dept. of Computer Science & Engineering, Muthiah Polytechnic College
[2] Dept. of Computer Science & Engineering, Annamalai University
[3] Centre of Advanced Studies in Linguistics, Annamalai University
Annamalainagar. Tamil Nadu, India

**ABSTRACT:**

*This paper presents machine learning solutions to a practical problem of Natural Language Generation (NLG), particularly the word formation in agglutinative languages like Tamil, in a supervised manner. The morphological generator is an important component of Natural Language Processing in Artificial Intelligence. It generates word forms given a root and affixes. The morphophonemic changes like addition, deletion, alternation etc., occur when two or more morphemes or words joined together. The Sandhi rules should be explicitly specified in the rule based morphological analyzers and generators. In machine learning framework, these rules can be learned automatically by the system from the training samples and subsequently be applied for new inputs.*

*In this paper we proposed the machine learning models which learn the morphophonemic rules for noun declensions from the given training data. These models are trained to learn sandhi rules using various learning algorithms and the performance of those algorithms are presented. From this we conclude that machine learning of morphological processing such as word form generation can be successfully learned in a supervised manner, without explicit description of rules. The performance of Decision trees and Bayesian machine learning algorithms on noun declensions are discussed.*

## [1] INTRODUCTION

Most of the Indian languages are morphologically rich. Words in these languages are formed by combining meaningful constituents, called **morphemes**. There are free morphemes which can function independently as words, and bound morphemes appear only as parts of words, always in conjunction with a root and sometimes with other bound morphemes. The bound morphemes act as affixes to form inflectional and derivational categories. The way in which these constituents are organised in word-forms follows morphological rules. Morphemes may have varied forms in particular contexts, variations of that morphemes are called allomorphs.[1][13][18].

One important aspect of morphemes is that they change when grouped together. Actually, they influence each other phonologically. These changes are called **morphophonemic changes**. When suffixes are added with words, a change in the word, like addition of consonant or vowel takes place depending upon the nature of the suffix. The rules that make changes in words when adding suffixes are called **sandhi rules**. The word 'sandhi' is a Sanskrit word meaning 'meeting together'. Application of sandhi rules is necessary to correctly write and pronounce complex word forms [1]. Developing a rule based system to model these sandhi-changes involves many complex linguistic rules along with the list of affixes to be selected and concatenated to the root. The automatic language processing systems represent these rules in the form of Finite State Automata. [5][8][[14]

In this paper we present supervised machine learning algorithms such as Decision Trees and Bayesian classifiers which learn the rules of morphophonemic changes from the training data. The data are collected from the Tamil corpus. The machine learning algorithms are applied for the classification of character sequences in to one of the predefined inflectional classes.

The rest of the paper is organised as follows. Section 2 provides an overview of prior works addressing the Morphological generation and the application of machine learning techniques in language processing. Section 3 introduces the Tamil language and the general morphophonemic concepts related to this topic. Description of the machine learning techniques used in this paper is given in Section 4. The implementation and the results of the machine learning models are discussed in Section 5.

## [2] RELATED WORKS

One of the earlier rule based morphological analyser was developed for Tamil at CIIL [12]. A generic architecture for morphological generators of agglutinative languages using paradigms and allomorph table has been proposed [45], in which corpus is used to extract fully inflected word forms. In [5][14], they have discussed the algorithm for developing a rule based tagger for Tamil. The rule based morphological analyser; morphological generator [14] is available for download, on Tamil Virtual University portal. RCILTS for Tamil has developed the morphological analyser 'Atcharam' and the generator 'Atchayam'. Other rule based morphological analysers and generators for Tamil are discussed in [4][22][36][38][45][47]. In [3], authors have proposed Sequence Labeling Approach for morphological analysis. Machine learning approaches have been presented for for learning past tense of English verbs [40], Tamil document classification [32], Morphological analysis [8][46], Parts-of-speech tagger using Neural networks [2][10][30][33][42] and Decisoin trees[20][41].

The existing morphological analyzers and generators for Tamil divide the verbs and nouns into various paradigms, and input word stems are compared with the paradigm tables. The major components of those systems are: i) A lexicon of roots with category information. Ii) A list of inflections for each category. Iii) A set of morphophonemic rules and iv) A set of tables for projecting allomorphic variations.

Decision trees have been applied to a wide range of NLP problems including grapheme-to-phoneme conversion [17], part-of-speech tagging [41], tokenization [Palmer & Hearst 1997], parsing [19][20], morphological analysis [46] and language modelling [15]. Many of these methods directly apply decision trees as classifiers [4].

## [3] TAMIL LANGUAGE

# MACHINE LEARNING OF PHONOLOGICALLY CONDITIONED NOUN DECLENSIONS FOR TAMIL MORPHOLOGICAL GENERATORS

Tamil is an agglutinative and concatenative language, where morphemes are strung together to form long words. Tamil is a verb final language with SOV (Subject Object Verb) pattern. The verbs can have a long sequence of morphemes that express tense, mood and aspects as well as sense of assertion, negation, interrogation, reflection, emphasis, etc. Nouns take plural marker to make plural forms, and they decline to different inflected forms by taking case markers [1][16].

In modern Tamil, all lexical or root morphemes can be distinguished into four groups: two major groups of nominal and verbal roots, which comprise almost all roots occurring in modern Tamil, and two minor groups of adjectival roots, and adverbial roots [18]. The verbs and nouns are more productive in the sense that they can produce more word forms. Nouns can also be derived from verbal roots. In the Tamil corpus, the total number of nouns is very much higher than the verbs.

## [3.1] Morphophonemics

Morphophonemics is the study which deals with the phonemic variations of a single morpheme in the given environment. It deals with all kinds of changes like addition, deletion or change etc., when two or more morphemes occur together when they stand in conjunction. It is undeniable that the morphological aspect is indispensable for making sandhi rules. **Internal sandhi** occurs when two morphemes are added with one another and **external sandhi** occurs between two free-forms or two grammatical forms. The internal sandhi occurs in the morphological structure, that is, it relates to the sandhi of the morphemes within a word. So it is called "*morphological sandhi*". The latter one occurs in the syntactic structure of a language. So, it is called compound or phrase or "*syntactical sandhi*". A morphophonemic rule has the form of a phonological rule, but is restricted to a particular morphological environment [39].

An example of a morphophonological alternation in English is provided by the plural morpheme, written as "-s" or "-es". Its pronunciation alternates between [s], [z], and [ɪz], as in *cats*, *dogs*, and *horses* respectively. A purely phonological analysis would likely assign to these three endings the phonemic representations /s/, /z/, /ɪz/. On a morphophonological level, however, they may all be considered to be forms of the underlying object //z//, which is a *morphophoneme* [6].

Affixes can be realized in many ways. For example, past tense is realized with **î¢, î¢î¢, ï¢î ¢, ø¢,** and **ì¢** and plural is realized with **è÷¢, è¢è÷¢, é¢è÷¢**. These alternations are conditioned by the phonological properties of the surrounding sound segments of the affixes. It is quite common for the same morpheme to be realized in different ways. This variation in form is known as allomorphy, and the variants of one morpheme are called allomorphs. Variation is conditioned by the environment in which the morpheme occurs; it can be phonologically, or grammatically or lexically conditioned.

When adding case suffixes to nouns, nouns undergo sandhi changes. For example, when a word that ends in a vowel and a suffix that begins in a vowel are added together, a glide - either **ò¢** or **õ¢** - is inserted in between. Selection of either **ò¢ or õ¢** is determined based upon whether the final vowel is

1) One of the 'front' vowels **Þ, ß, â, ã, ä**   or
2) One of the 'back' vowels **Ü, Ý, å, æ, à, á** respectively.

Front vowels take the glide ò¢ and back vowels take the glide õ¢. Some nouns ending with consonants, with a syllable structure CVC where the vowel is short is added with any suffix, the final consonant is doubled. The morphophonemic processes that are involved in Tamil morphology are assimilation, insertion, deletion and germination. Words with the syllable structure CVC, nouns ending in Üñ¢, ´ and Á undergo a number of different changes when a case suffix is added [27] [39].

### [3.2] Noun declension

In Tamil, nouns can be inflected for both number and case, the plural suffix is first added to the noun stem, optionally followed by the euphonic increment *–in/-an*, and then the case suffix is added. An inflected noun form may be the realisation of three morphemes, as given in the following representation

Noun stem + [ Plural suffix ] + [Euphonic increment] + [Case suffix]

When a noun is inflected for case only, a case suffix is added either directly to the stem or to the oblique stem in case the respective noun has such a stem. When a noun is inflected for both number and case, the plural suffix is first added to the noun stem, optionally followed by the euphonic increment –in, and then the case suffix is added. When different types of suffixes are added to the noun, various morphophonemic rules operate [18]. The list of suffixes for plural and various cases are shown in Table 1.

| Category | Suffixes |
|---|---|
| Plural | -è÷¢ |
| Euphonic increment | -Þù¢ / Üù¢ |
| Case Suffixes | |
| Nominative | NULL |
| Accusative | -ä |
| Instrumental | -Ýô¢ |
| Dative | -àè¢°/è¢° |
| Locative | -Þô¢ |
| Ablative | -Þô¤¼ï¢¶ |
| Sociative | -àìù¢/æ´ |
| Genitive | -Þù¢/ÞÂ¬ìò |

Table 1. List of Inflectional Suffixes

### [3.3] Case Marking

Case is essentially a system of marking dependent nouns for the type of relationship they bear to their heads. Traditionally, the term refers to inflectional marking, and, typically, case marks the relationship of a noun to a verb at the clause level or of a noun to a postposition, or another noun at the phrase level. The nominative is the citation form and is used for the subject of a clause. The accusative is used for the direct object and the dative for the indirect object (the recipient of a verb of giving). The genitive expresses the possessor, and the sociative expresses the notion of 'being in the company of'. The locative expresses location, and the instrumental expresses the instrument, as in 'cut with a knife' and the agent of the passive. The ablative expresses 'from'.

The following table 2. shows the types of sandhi changes for the given noun stems when case suffixes and plural suffixes are added.

| Category | Suffixes | ò¢–inserted | õ¢-inserted | Consonant Doubling | î¢î¢ inserted (-ñ¢) |
|---|---|---|---|---|---|
| Nominative | NULL | ð® | ð² | èí¢ | ðìñ¢ |
| Accusative | -ä | ð®¬ò | ð²¬õ | èí¢¬í | ðîî¢¬î |
| Instrumental | -Ýô¢ | ð®ò£ô¢ | ð²õ£ô¢ | èí¢í£ô¢ | ðîî¢î£ô¢ |
| Dative | -àè¢° | ð®è¢° | ð²¾è¢° | èí¢µè¢° | ðîî¢¶è¢° |
| Locative | -Þô¢ | ð®ò¤ô¢ | ð²õ¤ô¢ | èí¢í¤ô¢ | ðîî¢î¤ô¢ |
| Ablative | -Þô¢¼ï¢¶ | ð®ò¤ô¢¼ï¢¶ | ð²õ¤ô¢¼ï¢¶ | èí¢í¤ô¢¼ï¢¶ | ðîî¢î¤ô¢¼ï¢¶ |
| Sociative | -àìù¢/ æ´ | ð®»ìù¢ | ð²¾ìù¢ | èí¢µìù¢ | ðîî¢¶ìù¢ |
| Genitive | Þù¢/ÞÂ¬ìò | ð®ò¤Â¬ìò | ð²õ¤Â¬ìò | èí¢í¤Â¬ìò | ðîî¢î¤Â¬ìò |
| Plural | -è÷¢ | ð®è÷¢ | ð²è¢è÷¢ | èí¢è÷¢ | ðìé¢è÷¢ |

**Table 2. Various noun inflections and their plural forms for Tamil (Uninflected forms of nouns are given in Nominative case)**

## [4] MACHINE LEARNING

Machine learning is one of the important research and application areas of artificial intelligence (AI). Machine learning is concerned with acquiring knowledge from an environment in a computational manner. Machine learning is the capability of a computer to learn from experience (training data) and to extract knowledge from examples. A successful learner should be able to make general conclusions about the data it is trained on. This allows it to act appropriately in new situations. Many machine learning techniques have been applied to NLP tasks.

In the recent years, the application of machine learning based techniques to language learning and acquisition problems has been the focus of increasing attention in the NLP community. The problems in the natural language understanding can be recast as classification problems, a generic type of problem in the AI area. Machine learning methods include several symbolic inductive learning paradigms such as instance based learning, decision trees, threshold linear separators, inductive logic, unsupervised clustering, etc.; and also a number of sub-symbolic and connectionist approaches, such as neural networks and genetic algorithms [2][7][10][19].

Most natural language processing (NLP) tasks require the mapping of one level of representation to another. For example, in text to speech systems, the spelling representation of words is translated to a corresponding phonetic representation; in part-of-speech (POS) tagging, the words of a sentence are translated into their contextually appropriate POS tags. All these types of NLP tasks can be formulated as a *classification task*, and are therefore appropriate problems for machine learning methods. Classification-based learning starts from a set of instances (examples), each consisting of a set of input features (a feature vector) and an output class. In general, machine learning algorithms can be classified as supervised learning and unsupervised learning. **Supervised learning** uses a set of inputs for which the appropriate (i.e., desired) outputs are known [11][24].

### [4.1] Decision Trees

Decision tree classifier is one of the techniques for solving classification problems [24][25]. Decision trees are easy to create, to understand, and to apply, and they are quite accurate. They are easy to interpret and can be re-represented as *if-then-else-rules*. Decision trees are learned from training data. Each data item consists of a set of features describing an object and the class of the object. Decision trees are recursively built beginning with the top most node by (i) computing the best test for the current node according to some splitting criterion, (ii) creating a sub node for each possible outcome of the test, and (iii) recursively expanding each sub node in the same way until a given stopping criterion is satisfied. Usually, the decision tree is simplified (pruned) in order to avoid over-fitting of the training data.

Quinlan's ID3 decision tree building algorithm and its variations such as C4.5 have become one of the most widely used symbolic learning techniques. Given a set of objects, ID3 produces a decision tree that attempts to classify all the given objects correctly. At each step, the algorithm finds the attribute that best divides the objects into the different classes by minimizing entropy. After all objects have been classified or all attributes have been used, the results can be represented by a decision tree or a set of production rules [28][29].

### [4.2] Bayesian Model

Bayesian model is one of the popular probabilistic models in pattern recognition research. This model is often used to classify different objects into predefined classes based on a set of features. A Bayesian model stores the probability of each class, the probability of each feature, and the probability of each feature given each class, based on the training data. When a new instance is encountered, it can be classified according to these probabilities. A Naïve Bayesian model, variation of the Bayesian model, assumes that all features are mutually independent within each class [21][48].

### [5] IMPLEMENTATION

### [5.1] Methodology

Input to the machine learning algorithm consists of a collection of training data, each having a tuple of values for a fixed set of attributes (or independent variables) and a class attribute (or dependent variable). An attribute is described as continuous or discrete according to whether its values are numeric or nominal. The class attribute is discrete.

In this paper, we present the training data as a sequence of nominal (character) values. The words of the corpus are converted into consonant and vowel sequence. The morphophonemic rules can be learned automatically from the examples given to the system. The characters to the left of the focus point (here, the sandhi position), and the characters to the right are given as input features. For modeling inflections, the stem and suffixes are the two components and are used as the left and right context of the sandhi position.

The experiment is conducted using the sample data collected from the Tamil corpus. The total of 35000 noun forms which are inflected with various case markers and plural forms. From which 4047 distinct noun stems are selected as training data. Different types of sandhi changes are assigned unique class numbers as shown in the Table 3. The class numbers indicate the type of change occurs when the case and plural markers are added with the noun stem.

| Class | Sample data |
|---|---|
| 1 - ò¢–insertion | ð® + ä → ð®¬ò |
| 2 - õ¢-insertion | ð² + ä → ð²¬õ |
| 3  Consonant Doubling | èô¢ + ä → èô¢¬ô |
| 4 -î¢î¢ inserted (-ñ¢ removed) | ðìñ¢ + ä → ðìî¢¬î |
| 5 - à deletion | õ¤ø° + ä → õ¤ø¬è |
| 6 - à deletion, Consonant Doubling | è£´ + ä → è£ì¢¬ì |
| 7 No change (Normal) | è£ô¢ + ä → è£¬ô |
| **Plural forms** | |
| 7  Plural form 1(No change)  ~è÷¢ | ð® + è÷¢ =ð®è÷¢ |
| 8  Plural form 2    ~é¢è÷¢ | ðìñ¢ + è÷¢= ðìé¢è÷¢ |
| 9  Plural form 3    ~è¢è÷¢ | ð² + è÷¢ = ð²è¢è÷¢ |
| 10 Plural form 4(Consonant change ô¢-ø¢)   ~ø¢è÷¢ | èô¢ + è÷¢ = èøè¢è÷¢ |
| 11   Plural form 5 (Consonant change ÷¢-ì¢)  ~ì¢è÷¢ | ï£÷¢ + è÷¢ = ï£ìè¢è÷¢ |

Table 3. Class numbers assigned for the morphophonemic changes

[5.2] Feature representation

Different types of sandhi changes take place on different stem and suffix combinations. Training data is collected from the corpus for the above classes. The class numbers represent the type of sandhi changes. Maximum of 10 characters from the end of stem and 5 characters from suffix are used as nominal (symbolic) features. The symbol X in the feature value indicates a blank character. The class number is given as the output during training. The table 4, shows the feature vector for è£´ + ä and ðìñ¢ + è÷¢ which belongs to output categories 6 and 8 respectively.

| Stem features (10) | Suffix Features | Class |
|---|---|---|

| 1 | 2 | 3 | 4 | 5 | 6 | 7 | 8 | 9 | 10 | 11 | 12 | 13 | 14 | 15 | 16 |
|---|---|---|---|---|---|---|---|---|---|---|---|---|---|---|---|
| X | X | X | X | X | X | è¢ | Ý | ì¢ | à | | ä | X | X | X | X | 6 |
| X | X | X | X | X | | ð¢ | Ü | ì¢ | Ü | ñ¢ | è¢ | Ü | ÷¢ | X | X | 8 |

Table 4. Feature vector with input features and output class (Model I)

| Stem Features (5) | | | | | | Suffix features | | | | Class |
|---|---|---|---|---|---|---|---|---|---|---|
| 1 | 2 | 3 | 4 | 5 | | 6 | 7 | 8 | 9 | 10 | 11 |
| X | X | è¢ | Ý | ô¢ | | ä | X | X | X | X | 7 |
| ð¢ | Ý | ì¢ | Ü | ñ¢ | | è¢ | Ü | ÷¢ | X | X | 8 |

Table 5. Feature vector with input features and output class (Model II)

### [5.3] Data sets

The morphophonemic variations due to phonological conditioning can be captured from few characters near the boundary (Sandhi position). To test this hypothesis, we changed the length of the context and prepared two different data sets. For the first data set (Model I), 10 character of the stem are used as left context. For the second experiment (Model II), only the right most 5 characters of the stem are used as left context. The learning algorithms are independently tested on these two sets of data.

### [5.4] Experimental Results

The training data of Model-I and Model-II in the specified format with different input attributes and single output attribute are supplied to the decision tree and Bayes classifiers. The classifiers generalize from the training data. The machine learning algorithms were tested on a Java based WEKA open source machine learning tool. The Weka machine learning tool [48] which has become a standard benchmarking tool in the machine learning community in recent years, was used to build models. The classifier models in this environment accept nominal values for the features. The classification is performed on various tree classifiers and Bayes classifiers with default parameter settings and a 10 fold cross validation.

The algorithms used are listed below.
    1.     BayesNet     (B_Net)
    2.     NaiveBayes     (N_B)
    3.     Bayes Averaged, one-dependence estimators (B_AODE)
    4.     Basic divide-and-conquer decision tree algorithm (ID3)
    5.     C4.5 decision tree learner ( J48)
    6.     Random Tree (R_Tree)
    7.     RandomForest (R_Forest)

The performance measures for these algorithms are : Correctly classified Instances (CCI), Incorrectly classified instances (ICI), Kappa Statistics (KS), mean absolute error (MAE), relative squared error (RAE), root mean-squared error (RMSE) and root relative squared error (RRSE).

| Measures | B_Net | N_B | B_AODE | Id3 | J48 | R_tree | R_Forest |
|---|---|---|---|---|---|---|---|
| CCI Numbers % | 3947 97.529 | 3939 97.3314 | 3913 96.6889 | 3890 96.1206 | 4001 98.8634 | 3361 83.0492 | 3987 98.5174 |
| ICI Numbers % | 100 2.471 | 108 2.6686 | 134 3.3111 | 157 3.8794 | 46 1.1366 | 686 16.9508 | 60 1.4826 |
| KS | 0.9674 | 0.9647 | 0.956 | 0.9875 | 0.9849 | 0.7737 | 0.9804 |
| MAE | 0.0101 | 0.0113 | 0.0184 | 0.0028 | 0.0056 | 0.0503 | 0.0189 |
| RMSE | 0.0741 | 0.0755 | 0.0875 | 0.0518 | 0.0548 | 0.1975 | 0.0721 |
| RAE | 4.6943 | 5.207 | 8.5093 | 1.3566 | 2.5973 | 23.2613 | 8.7639 |
| RRSE | 22.532 | 22.9651 | 26.623 | 16.0304 | 16.6873 | 60.0963 | 21.9298 |

Table 6. Performance of classifiers on Model-I Data set (With 10 Characters)

| Measures | B_Net | N_B | B_AODE | Id3 | J48 | R_tree | R_Forest |
|---|---|---|---|---|---|---|---|
| CCI Numbers % | 3982 98.3939 | 3974 98.1962 | 3952 97.6526 | 3959 97.8255 | 4001 98.8634 | 3929 97.0843 | 3998 98.7892 |
| ICI Numbers % | 65 1.6061 | 73 1.8038 | 95 2.3474 | 88 2.1744 | 46 1.1366 | 118 2.9157 | 49 1.2108 |
| KS | 0.9787 | 0.9761 | 0.9688 | 0.9874 | 0.9849 | 0.9613 | 0.984 |
| MAE | 0.0085 | 0.0097 | 0.0148 | 0.0036 | 0.0056 | 0.0093 | 0.0046 |
| RMSE | 0.0629 | 0.0662 | 0.0721 | 0.0479 | 0.0548 | 0.0813 | 0.0523 |
| RAE | 3.9206 | 4.5043 | 6.8369 | 1.6655 | 2.5973 | 4.2897 | 2.1237 |
| RRSE | 19.1305 | 20.1311 | 21.9353 | 14.6789 | 16.6873 | 24.7303 | 15.8976 |

Table 7. Performance of classifiers on Model-II Data set (With 5 characters)

This shows that, phonological conditioning of sandhi rules can be captured from fewer characters at the boundary without performance degradation. The lower performance of the learning algorithms on data set having more features (10 stem features) is due to more empty characters on the stems. The overall performance of the machine learning algorithms on learning inflections from the phonologically conditioned input features shows that the grammar rules of sandhi changes can be captured easily from the training data.

**[6] CONCLUSION**

The sandhi rules are important in word formation. An automatic sandhi checker or sandhi generator can be used as a critical component in morphological analyser, morphological generator and text-to-speech synthesizer. In this paper we studied the performance of machine learning models which classify sandhi changes on noun inflections. Tamil grammar recognizes different kinds of conditioning for the sandhi changes like phonological, grammatical and syntactic relations. Even within the grammatical relation and syntactic relation, the phonological condition may operate. In this paper only the characters present in the noun stem and the suffixes are used as features, without considering any grammatical or syntactic relations between the two components. Additional syntactic information can be used as features for a generalised word formation tasks. These models can also be applied for verb conjugations and other morphological processing applications for agglutinative languages. Machine learning techniques capture the inherent patterns in the language and are suitable for language learning applications.

MACHINE LEARNING OF PHONOLOGICALLY CONDITIONED NOUN DECLENSIONS FOR TAMIL MORPHOLOGICAL GENERATORS

REFERENCES

[1]    Agesthialingam.S, Tamil Mozhi Amaippial, Annamalainagar, (2002)
[2]    Ahmed, S. Bapi Raju, P.V.S. Chandrasekhar, M. Krishna Prasad, Application of Multilayer Perceptron Network for Tagging Parts-of-Speech. Proceedings of the Language Engineering Conference (LEC'02) (2002), IEEE Computer Society: 57-63.
[3]    Anand Kumar M, Dhanalakshmi V, Soman K.P and Rajendran S, A Sequence Labeling Approach to Morphological Analyzer for Tamil Language, (IJCSE) International Journal on Computer Science and Engineering Vol. 02, No. 06, (2010),
[4]    Anandan.P, Rajani Parthasarathy, Geetha, T.V, Morphological Generator for Tamil, in Tamil Internet 2001 Conference Proceedings, Malaysia.
[5]    Arulmozhi.P, Sobha.L and Kumara Shanmugam. B, Parts of Speech Tagger for Tamil. Symposium on Indian Morphology. Phonology & Language Engineering, (2004), IIT Kharagpur. :55-57.
[6]    Booij.G, The Grammar of Words: An Introduction to Linguistic Morphology. (2004), Oxford University Press.
[7]    Daelemans.W, & van den Bosch.A, Memory-Based Language Processing. Cambridge: Cambridge University Press. (2005).
[8]    Dhanalakshmi V. et.al, Morphological Analyzer for Agglutinative Languages Using Machine Learning Approaches, Proceedings of International Conference on Advances in Recent Technologies in Communication and Computing,(2009)
[9]    Dietterich T.G.(1997), Machine learning research: Four Current Directions, AI Magazine, 18(4):97-136
[10]   Elman J.L,(1990) Finding structure in Time, Cognitive Science 14 :179-211.
[11]   Etham Alpaydin, Introduction to Machine Learning, The MIT Press Cambridge, Massachusetts, London, England, (2004).
[12]   Ganesan.M, A scheme for Grammatical Tagging of Corpora, In B.B.Rajapurohit (Ed.) Technology and Languages. Mysore: CIIL, (1994).
[13]   Ganesan.M, Computational Grammar of Tamil, In R.Hari (Ed.) Word Structure in Dravidian. Kuppam: Dravidian University. (2003).
[14]   Ganesan.M, Tamil Morphology: A Computational Model. New perspectives in Linguistics , Ed. Shanmugam.C et.al,(2010).
[15]   Jurafsky.D, and Martin.J, "Speech and Language Processing", Prentice Hall, 2000.
[16]   Karthikeyan.A, Compound verbs in Tamil, Unpublished Ph.D Thesis, Annamalai University,1983
[17]   Kienappel. A. K, & Kneser .R, Designing very compact decision trees for graphem-to-phoneme transcription. Proceedings of the Eurospeech Conference 1911–14,2001.
[18]   Lehman. T, A grammar of Modern Tamil. Pondicherry: Pondicherry Institute of Linguistics and Culture.1993.
[19]   Magerman.D.M, Statistical decision-tree models for parsing. In Meeting of the Association for Computational Linguistics, pages 276–283. (1995).
[20]   Marquez.L,et al.Automatically acquiring a Language model for POS using Decision Trees. In Proceedings of the second Conference on Reccent Advances in Natural Language Processing, RANLP, pages 27-34.
[21]   Matthew. G, Et al, A Bayesian Model for Morpheme and Paradigm Identification, In Proceedings of the 39th Annual Meeting of the ACL, pages:482-490. ( 2001).

[22] Menaka .S, Vijay Sundar Ram and Sobha Lalitha Devi, Morphological Generator for Tamil, In "Morphological Analysers and Generators", (ed.) Mona Parakh, LDC-IL, Mysore, pp. 82 –96, 2010.

[23] Michael C. Frank, Ichinco.D, and Tenenbaum.J.B., Principles of generalization for learning sequential structure in language. In Proceedings of the 30th Annual Meeting of the Cognitive Science Society, page 763-768. 2008.

[24] Mitchell.T, Decision Tree Learning, in Machine Learning, The McGraw-Hill Companies, Inc., pp. 52-78. 1997.

[25] Murthy S.K., Automatic construction of decision trees from data: A multi disciplinary survey, Data Mining and Knowledge Discovery, Vol.2, pp 345-389,1998

[26] Nuhman.M.A, Basic Tamil Grammar, University of Peradeniya ,Sri Lanka. (1999).

[27] Pon Kothandaraman , A Grammar of contemporary Literary Tamil, International Institute of Tamil Studies, Chennai.(1997).

[28] Quinlan, Ross.J, C4.5: Programs for machine learning. San Francisco:Morgan Kaufmann. (1993).

[29] Quinlan, Ross.J, , Induction of decision trees. Machine Learning 1:81–106.

[30] Rajan .K, et.al., Applications of Neural Networks for Tamil Studies, International seminar on Tamil Computing, Madras University, Chennai. (2002),

[31] Rajan .K, et.al., Corpus Analysis and Tagging for Tamil, Symposium on Translation Support Systems, I.I.T. Kanpur. ,(2002)

[32] Rajan.K, Ramalingam.V, Ganesan.M, Palanivel.S, Automatic classification of Tamil documents using vector space model and artificial neural network, In Elsevier,Expert Systems with Applications, 36, 10914–10918. (2009).

[33] Rajan.K, Ramalingam.V, Ganesan.M, Computational approaches for learning inflections in Tamil, Tamil Internet Conference, pg.183-189.(2010).

[34] Rajan.K, Ramalingam.V, Ganesan.M, Application of Genetic Algorithm for morphological segmentation, In Proceedings of National Seminar on 'New Perspectives in Applied Linguistics', CAS Linguistics, Annamalai University, 192-195, December 2011.

[35] Rajan.K, Ramalingam.V, Ganesan.M, Computational models for Morphotactics, In Proceedings of National Seminar on Computational Linguistics and Language Technology,45-49, CAS Linguistics, Annamalai University. (2011).

[36] Rajan.K.,et.al, Corpus analysis tools for Tamil, Tamil Internet 2003, Chennai.

[37] Raymond J. Mooney and Mary Elaine Califf, Learning the Past Tense of English Verbs Using Inductive Logic Programming, in Symbolic, Connectionist, and Statistical Approaches to Learning for Natural Language Processing, Springer Verilag, ,(1996)

[38] Rekha R U, Anand kumar M, Dhanalakshmi V Soman K P , Morphological Generator for Tamil A new data driven approach , pg-141-145.

[39] Renganathan.D, Sandhi in Modern Tamil, Unpublished Ph.D Thesis, Annamalai University, ( 1983).

[40] Rumelhart, D. E., & McClelland, J. L. On learning the past tense of English verbs. In Parallel distributed processing: Explorations in the microstructure of cognition (Vol. 2). Cambridge,MA: MIT Press. (1986).

[41] Schmid.H, Probabilistic part-of-speech tagging using decision trees, In New Methods in Language Processing, Jones D.B.(Ed.), UCL Press. (1997).

[42] Schmid.H, Parts-of-speech tagging with neural network, In COLING 94: The 15th International conference in Computational Linguistics, Kyoto, Japan. (1994).

[43] Sproat, Richard, Morphology and Computation. Cambridge, MA: MIT Press.1992.

[44] Thomas Lehman, A grammar of modern Tamil, Pondicherry Institute of Linguistices and culture. 1989.

[45] Uma Maheshwar Rao G, Parameshwari K: CALTS, University of Hyderabad, 'On the description of morphological data for morphological analyzers and generators: A case of Telugu, Tamil and Kannada'.

[46] Van den Bosch, A., Walter Daelemans, & T.Weijters, Morphological analysis as classification: an inductive learning approach. Proceedings of NEML. (1996).

[47] Viswanathan.S, Ramesh Kumar.S, Kumara Shanmugham.B, Arulmozi.B and Vijay Shanker.K, A Tamil Morphological Analyzer, ICON-2003, pp. 31-39.2003

[48] Witten .I.H & Frank.E,Data Mining: Practical machine learning tools and techniques, Morgan Kaufmann. 2005.